\documentclass[runningheads]{llncs}
\usepackage{graphicx}

\usepackage{tikz}
\usepackage{comment}
\usepackage{amsmath,amssymb} 
\usepackage{color}
\usepackage{graphicx}
\usepackage{multirow}
\usepackage{algorithm}
\usepackage{algorithmicx}
\usepackage{algpseudocode}
\usepackage[resetlabels]{multibib}
\newcites{Appx}{References}

\usepackage[accsupp]{axessibility}  
\usepackage[misc]{ifsym}

\newcommand{\tran}{{^{\mkern-1.5mu\mathsf{T}}}}
\newcommand{\ens}{\mathrm{ens}}
\newcommand{\com}{\mathrm{com}}
\newcommand{\SFDA}{\mathrm{SFDA}}
\newcommand*{\defeq}{\stackrel{\text{def}}{=}}
 
\definecolor{candypink}{rgb}{0.89, 0.44, 0.48}

\usepackage[accsupp]{axessibility}  

\begin{document}
\pagestyle{headings}
\mainmatter
\def\ECCVSubNumber{100}  

\title{Not All Models Are Equal: Predicting Model Transferability in a Self-challenging Fisher Space}

\titlerunning{Predicting Model Transferability in a Self-challenging Fisher Space}
%
\author{Wenqi Shao\inst{1,2}$^\star$, Xun Zhao\inst{2}, Yixiao Ge\inst{2}$^\star$, Zhaoyang Zhang\inst{1}, \\Lei Yang\inst{3}, Xiaogang Wang\inst{1}, Ying Shan\inst{2}, Ping Luo\inst{4}}
\authorrunning{W. Shao et al.}
%

\institute{{\textsuperscript{1} The Chinese University of Hong Kong ~~ \textsuperscript{2} ARC Lab, Tencent PCG} \\
\textsuperscript{3} Applied Model Center, Tencent PCG ~~
\textsuperscript{4} The University of Hong Kong \\
\email{weqish@link.cuhk.edu.hk} ~~ \email{pluo.lhi@gmail.com}
\email{\{yixiaoge,yingsshan\}@tencent.com} \\
$^\star$ Corresponding authors: Wenqi Shao, Yixiao Ge}


\maketitle

\begin{abstract}
This paper addresses an important problem of ranking the pre-trained deep neural networks and screening the most transferable ones for downstream tasks. It is challenging because the ground-truth model ranking for each task can only be generated by fine-tuning the pre-trained models on the target dataset, which is brute-force and computationally expensive. Recent advanced methods proposed several lightweight transferability metrics to predict the fine-tuning results. However, these approaches only capture static representations but neglect the fine-tuning dynamics. 
To this end, this paper proposes a new transferability metric, called \textbf{S}elf-challenging \textbf{F}isher \textbf{D}iscriminant \textbf{A}nalysis (\textbf{SFDA}), which has many appealing benefits that existing works do not have. First, SFDA can embed the static features into a Fisher space and refine them for better separability between classes. Second, SFDA uses a self-challenging mechanism to
encourage different pre-trained models to differentiate on hard examples. Third, SFDA can easily select multiple pre-trained models for the model ensemble. Extensive experiments on $33$ pre-trained models of $11$ downstream tasks show that SFDA is efficient, effective, and robust when measuring the transferability of pre-trained models. For instance, compared with the state-of-the-art method NLEEP, SFDA demonstrates an average of $59.1$\% gain while bringing $22.5$x speedup in wall-clock time. The code will be available at \url{https://github.com/TencentARC/SFDA}.


\end{abstract}

\section{Introduction}
Due to the wide applications of DNNs \cite{he2016identity,he2016deep,dosovitskiy2020image}, an increasing number of pre-trained models are produced by training on different source datasets~ (\textit{e.g.}, ImageNet \cite{krizhevsky2012imagenet}), with different learning strategies~(\textit{e.g.}, self-supervised learning \cite{he2020momentum,grill2020bootstrap}). These pre-trained models are crucial in providing a good warm start for fine-tuning on target tasks in transfer learning. An interesting question naturally emerges: given the numerous pre-trained models, how can we properly and quickly rank the models thereby selecting the best ones for a certain target task. 
%
\begin{figure}[t]
\centering
\includegraphics[scale=0.41]{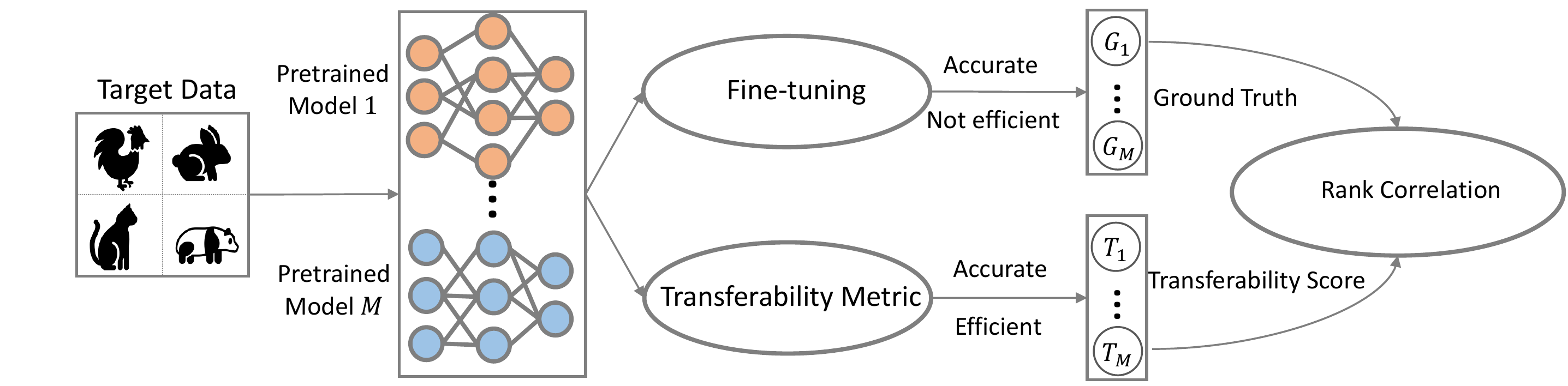}
\caption{The diagram of the task of ranking pre-trained models.
The ground truth of the problem is to fine-tune pre-trained models and collect fine-tuning accuracy $\{G_m\}_{m=1}^M$. However, it is inefficient to enumerate all models in a fine-tuning procedure with a hyper-parameters sweep. An efficient approach is to adopt several lightweight transferability metrics to predict the fine-tuning results. Each metric produces transferability scores $\{T_m\}_{m=1}^M$ for all pre-trained models. An evaluation based on rank correlation such as Kendall's tau assesses the transferability metric.}
\label{fig:fig1}
\end{figure}

Ranking pre-trained models is critical for transfer learning from a practical perspective.
Due to the lack of sufficient labelled data in various domains (\textit{e.g.}, object detection), a pre-trained model with expert knowledge is highly desired for initialization of downstream tasks in these domains.
Ranking pre-trained models is also challenging, as the optimal pre-trained model is usually task-specific \cite{li2021ranking}. 
The pre-trained model ranking method should be efficient and generic enough so that the best model can be quickly selected for downstream tasks.

The ground-truth ranking of pre-trained models is obtained by fine-tuning pre-trained models on the target dataset and ranking them by the test accuracies, as shown in Fig.\ref{fig:fig1}.
%
Brute-force fine-tuning is obviously time-consuming and computationally expensive.
For instance, fine-tuning one pre-trained model on one target dataset usually costs several GPU hours, selecting the optimal pre-trained model out of 10 models on 10 target tasks would cost many GPU days.
Recent works propose several lightweight transferability metrics such as LEEP \cite{nguyen2020leep} and LogME \cite{you2021logme} to substitute cumbersome fine-tuning procedure. 
Although these metrics are efficient to obtain, they simply measure the quality of pre-trained models by their static features, preventing themselves from characterizing the dynamics of the fine-tuning process, as shown in Fig.\ref{fig:fig2}.

In this paper, we propose a new transferability metric for ranking pre-trained models, namely \textbf{S}elf-challenging \textbf{F}isher \textbf{D}iscriminant \textbf{A}nalysis (\textbf{SFDA}). 
To approximate the fine-tuning process, SFDA captures two significant features of fine-tuning, i.e. classes separability and discrimination on hard examples, as shown in Fig.\ref{fig:fig3}. Towards this goal, we equip SFDA with a regularized FDA (Reg-FDA) module and a self-challenging mechanism. 
On the one hand, the Reg-FDA module projects the static features to a Fisher space where the updated features present better class separability. On the other hand, the self-challenging mechanism challenges SFDA by increasing the difficulty in separating classes, encouraging different pre-trained models to discriminate on hard examples.
Due to the above designs, SFDA behaves more like a fine-tuning process than prior arts \cite{li2021ranking,you2021logme,nguyen2020leep}, characterizing itself as an effective, efficient and robust transferability assessment method.
Moreover, SFDA can be naturally extended to multiple pre-trained model ensembles selection because features in the Fisher space for different pre-trained models have homogeneous dimensionality. Such homogeneity makes it possible to consider the complementarity between models.


The \textbf{contributions} of this work are three-fold. (1) We propose a new transferability metric named Self-challenging Fisher Discriminant Analysis (SFDA) which is effective, efficient and robust for ranking pre-trained models. (2) SFDA can be naturally extended to multiple pre-trained model ensembles selection. (3) Extensive experiments on $33$ pre-trained models produced by different types of architectures and training strategies over $11$ downstream classification tasks demonstrate the effectiveness of SFDA. For example, SFDA shows an average of $59.1$\% gain in the rank correlation with the ground truth fine-tuning accuracy while bringing 22.5x speedup in wall-clock time compared with previous state-of-the-art method NLEEP.

\section{Related Work}
\textbf{Transfer learning.} Transfer learning \cite{thrun1998learning,tan2018survey} has been extensively investigated in the form of different learning tasks such as domain adaptation \cite{wang2018deep,long2015learning} and task transfer learning \cite{zamir2018taskonomy}. 
In deep learning, transfer learning mainly comes in the form of inductive transfer, which usually refers to the paradigm of adapting the pre-trained models to target tasks \cite{yosinski2014transferable,kornblith2019better}. With the rapid development of deep learning in applications, numerous pre-trained models have been stored, such as HuggingFace Transformers \cite{wolf2019huggingface}. This paper focuses on ranking pre-trained models and selecting the best performing model for the target task. Unlike task transfer learning, where the prior knowledge in the source domain is known, ranking pre-trained models only has access to the pre-trained model itself. Hence, the method for ranking pre-trained models should be generic and efficient enough to apply to various pre-trained models and target tasks.

\noindent\textbf{Transferability of pre-trained models.} 
Measuring the transferability of pre-trained models on a target task has recently attracted much attention due to the broad practicability. LEEP \cite{nguyen2020leep} is a well-known early work studying this problem. It estimates the empirical joint probability of source and target label space. Hence, LEEP only works when the pre-trained model has a classification head that outputs source label probability. To pursue a more generic transferability method, NLEEP \cite{li2021ranking} generate a pseudo classification head by a Gaussian Mixture Model and LogME \cite{you2021logme} directly model the relationship between features extracted from the pre-trained models and their labels by marginalized likelihood. Although these metrics are easy to compute, they fail to characterize the dynamics of fine-tuning process, making the performance far from idealism. This paper proposes a new transferability metric that behaves more like fine-tuning by projecting features into a self-challenging Fisher space.

\section{Preliminaries}
In this section, we first present the setup, ground truth and evaluation protocol of the task of ranking pre-trained models and then introduce existing transferability metrics.

\noindent\textbf{Problem Setup.}
A target dataset with $N$ labeled data samples denoted as $\mathcal{T}=\{(x_n,y_n)\}_{n=1}^N$ and $M$ pre-trained models $\{\phi_m=(\theta_m, h_m)\}_{m=1}^M$ are given.
Each model $\phi_m$ consists of a feature extractor $\theta_m$ producing a $D$-dimension feature (i.e. $\hat{x}=\theta_m(x)\in\mathbb{R}^D$) and a classification head $h_m$ outputting label prediction probability given input $x$ \cite{he2016deep,dosovitskiy2020image}.
The task of pre-trained model ranking is to generate a score for each pre-trained model thereby the best model can be identified according to the ranking list.


\noindent\textbf{Fine-tuning as ground truth.} 
After fine-tuning all pre-trained models with hyper-parameters sweep on the target training dataset, the highest scores of evaluation metrics (e.g. test accuracy) are returned.
The fine-tuning accuracies of different pre-trained models are denoted as $\{G_m\}_{m=1}^M$,  which are recognized as the ground truth of pre-trained model ranking \cite{li2021ranking,you2021logme}. 
%

\begin{figure}[t]
\centering
\includegraphics[scale=0.39]{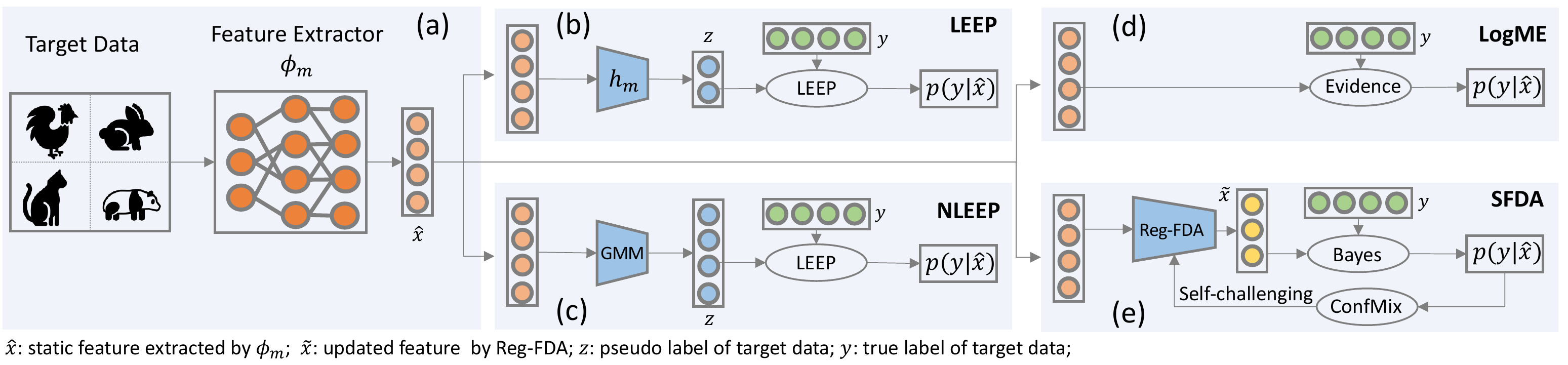}
\caption{Comparison of different transferability metrics including LEEP \cite{nguyen2020leep}, NLEEP \cite{li2021ranking}, LogME \cite{you2021logme} and our proposed SFDA. (a) shows a transferability metric rely on the static representation $\hat{x}$. (b-d) show that these metrics differentiate in modeling label prediction $p(y|\hat{x})$. Specifically, LEEP, NLEEP, and LogME calculate transferability score by static features $\hat{x}$, making it difficult to approximate fine-tuning as fine-tuning would update $\hat{x}$. Our proposed SFDA behaves more like fine-tuning in terms of classes separability and discrimination on hard examples, which are achieved by the module of Reg-FDA and a self-mechanism with the proposed ConfMix noise in (e).}
\label{fig:fig2}
\end{figure}

\noindent\textbf{Transferability metric.}
For each pre-trained model $\phi_m$, a transferability metric outputs a scalar score $T_m$ by
\begin{equation}\label{eq:metric-score}
T_m = \sum_{i=1}^N\log p(y_i|x_i; \theta_m, h_m)
\end{equation}
where $(x_i,y_i)$ denotes the $i$-th data point in target dataset $\mathcal{T}$.
A larger $T_m$ indicates that model $\phi_i$ can perform better on target $\mathcal{T}$. As we can see from Fig.\ref{fig:fig2}, transferability metrics differ in modelling label prediction probability $p(y_i|x_i; \theta_m, h_m)$. For instance, prior arts such as LEEP \cite{nguyen2020leep} obtain the probability based on static features $\hat{x}$, while ignoring that fine-tuning would update $\hat{x}$. Instead, our proposed SFDA is carefully designed to behave more like fine-tuning, as will be introduced in Sec.\ref{sec:SFDA}.





\noindent\textbf{Evaluation protocol.} 
Basically, if $\phi_i$ has higher classification accuracy than $\phi_j$ (i.e. $G_i > G_j$), $T_i > T_j$ is also expected. 
Hence, a rank-based correlation between $\{T_m\}_{m=1}^M$ and $\{G_m\}_{m=1}^M$ is competent to evaluate the effectiveness of transferability metrics. 
We use \textit{weighted Kendall's $\tau_w$} (detailed in Appendix Sec.A.1) by following the common practice \cite{you2021logme,li2021ranking}, where larger $\tau_w$ indicates better correlation and better transferability metric.

\section{Self-challenging Fisher Discriminant Analysis}\label{sec:SFDA}

We propose a Self-challenging Fisher Discriminant Analysis (SFDA) for the problem of pre-trained model ranking. SFDA  consists of two critical ingredients: a module of Regularized Fisher Discriminant Analysis (Reg-FDA) and a self-challenging mechanism motivated by two important observations in fine-tuning. As shown in Fig.\ref{fig:fig3}(a \& b), fine-tuning pre-trained model would update static features $\{\hat{x}_i\}_{i=1}^N$ for better classes separability. Moreover, Fig.\ref{fig:fig3}(d) show that different pre-trained models discriminate on hard examples during fine-tuning. Hence, the Reg-FDA and self-challenging mechanism are carefully designed to encourage classes separability and discrimination on hard examples. Algorithm 1 (see Appendix Sec.A.2) illustrates the whole pipeline of our proposed SFDA.

\textbf{Notation.} 
SFDA operates on static features $\hat{x}=\theta_m(x)$. 
Assume the target dataset $\mathcal{T}=\{(\hat{x}_n,y_n)\}_{n=1}^N$ has $C$ classes, we split it into classes. 
Then we have $\mathcal{T}=\{\hat{x}_n^{(1)}\}_{n=1}^{N_1} \cup \cdots \cup \{\hat{x}_n^{(C)}\}_{n=1}^{N_C}$, where $\hat{x}_n^{(c)}$ denotes the $n$-th instance of the $c$-th class and $N_c$ denote the sample size of the $c$-th class ($\sum_{c=1}^CN_c=N$).   


\subsection{Regularized FDA}\label{sec:Reg-FDA}

To approximate fine-tuning procedure, the primary thing is to transform $\{\hat{x}_i\}_{i=1}^N$ to a space such that the classes on target $\mathcal{T}$ are separated as much as possible as shown in Fig.3(b). 
Therefore, we propose Regularized Fisher Discriminant Analysis (Reg-FDA) for two reasons.
First, Reg-FDA promotes classes separability by inheriting the merits from conventional FDA \cite{mika1999fisher} that can maximize between scatter of classes and minimize within scatter of each class. 
%
Second, the optimization problem for Reg-FDA has a straightforward solution, which avoids gradient optimization and allows for efficient computation. 


\begin{figure}[t]
\centering
\includegraphics[scale=0.415]{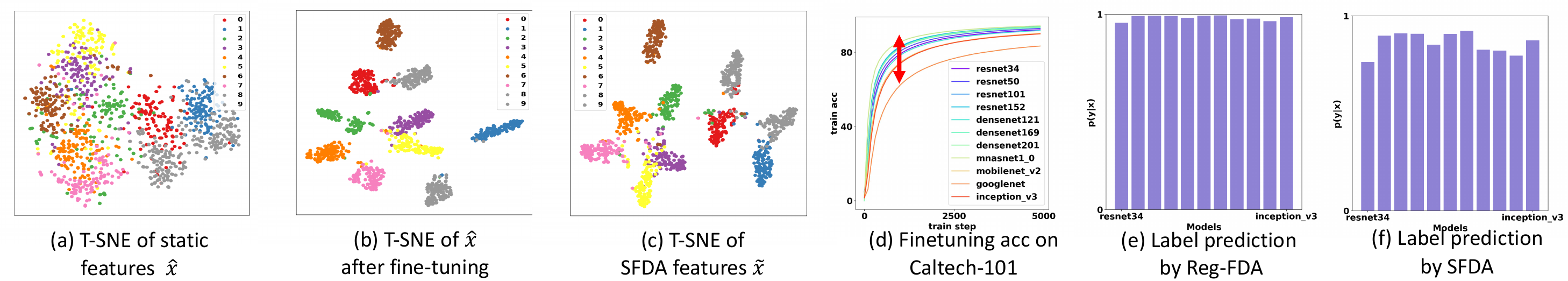}
\caption{(a-c) show that similar to fine-tuning our SFDA can update static features for better classes separability on the CIFAR-10 dataset. Deep models often fit easy examples in the early training stage while learning hard examples in the later training stage \cite{zhang2021understanding,arazo2019unsupervised}. Hence, (d) presents that different pre-trained models discriminate on hard examples. (e) and (f) show that SFDA can produce more discriminative label prediction results than Reg-FDA because the self-mechanism in SFDA encourages the different pre-trained models to discriminate on hard examples.}
\label{fig:fig3}
\end{figure}

\noindent\textbf{Formulation of Reg-FDA.} Reg-FDA defines a transformation that projects $\hat{x}\in\mathbb{R}^D$ into $\tilde{x}\in\mathbb{R}^{D'}$ by a projection matrix $U\in\mathbb{R}^{D\times D'}$. We have $\tilde{x} := U\tran\hat{x}$, where
\begin{equation}\label{eq:Reg-FDA}
U = \arg\max_U \frac{d_B(U)}{d_W(U)} \defeq\frac{|U\tran S_B U|}{|U\tran [(1-\lambda) S_W + \lambda I ]U|}
\end{equation}
In Eqn.(\ref{eq:Reg-FDA}), $d_B(U)=|U\tran S_B U|$ and $d_W(U)=|U\tran \tilde{S}_W U|$ represent between scatter of classes and within scatter of every class, where $\tilde{S}_W=(1-\lambda) S_W + \lambda I$ and $I$ is an identity matrix. Moreover, $S_B= \sum_{c=1}^CN_c(\mu_c - \mu)(\mu_c - \mu)\tran$ and $S_W=\sum_{c=1}^{C}\sum_{n=1}^{N_c}(\hat{x}_n^{(c)}-\mu_c)(\hat{x}_n^{(c)}-\mu_c)\tran$ are between and within scatter matrix respectively. Here $\mu=\sum_{n=1}^N\hat{x}_n$ and $\mu_c=\sum_{n=1}^{N_c}\hat{x}_n^{(c)}$ represent the mean of data and the mean of  $c$-th class respectively. $\lambda\in [0,1]$ in Eqn.(\ref{eq:Reg-FDA}) is a regularization coefficient used to trade off the inter-class separation and intra-class compactness. 
Reg-FDA degrades into FDA \cite{mika1999fisher} when $\lambda=0$.

\noindent\textbf{Adaptive regularization strength.} We treat $\lambda\in [0,1]$ as an regularization strength adaptive to different feature distribution. The motivation is that Reg-FDA can deal with diverse distribution of features $\{\hat{x}_i\}_{i=1}^N$ extracted from different pre-trained models when $\lambda$ is adaptively varied. For example, self-supervised ResNet-50 with Infomin has a larger within scatter of classes than its supervised counterpart on CIFAR-10 dataset as shown in Fig.5 of Appendix Sec.A.3, implying that ResNet-50 with Infomin needs stronger supervision on minimizing within scatter of every class for better classes separation. Motivated by this intuition 
we instantiate $\lambda$ as follows,
%
%
\begin{equation}\label{eq:lambda}
\lambda = \exp ^{-a\sigma(S_W)} \,\, \mathrm{where} \,\, \sigma(S_W) = \max_{u\in\mathbb{R}^D, \|u\|_2=1}u\tran S_W u.
\end{equation}
In Eqn.(\ref{eq:lambda}), $a$ is a positive constant and $\sigma(S_W)$ is the largest eigenvalue of $S_W$ by definition.
A larger $\sigma(S_W)$ indicates the larger within scatter. Hence, a smaller $\lambda$ should be used for the stronger supervision on minimizing within scatter.

\noindent\textbf{Efficient computation of Reg-FDA.} The optimization problem in Eqn.(\ref{eq:Reg-FDA}) can be solved efficiently as follows,
\begin{equation}\label{eq:solution-FDA}
S_B u_k = v_k \tilde{S}_W u_k
\end{equation}
Eqn.(\ref{eq:solution-FDA}) is a generalized eigenvalues problem with $v_k$ being the $k$-th eigenvalue and $u_k$ being the corresponding eigenvector. Note that $u_k$ constitutes the $k$-th column vector of the matrix $U$. The generalized eigenvalues problem can be efficiently solved by existing machine learning packages.
 Moreover, 
$\sigma(S_B)$ in Eqn.(\ref{eq:lambda}) is also easily obtained by the Iteration method, which only involves a matrix-vector product as provided in Appendix Sec.A.2.

\noindent\textbf{Classification by Bayes.} After obtaining projection matrix $U$, we then acquire updated feature representations  $\{\tilde{x}_n=U\tran \hat{x}_n\}_{n=1}^N$ which exhibits better class separability as shown in Fig.\ref{fig:fig3}(c). For each class, we assume $\tilde{x}_n^{(c)} \sim \mathcal{N}(U\tran\mu_c, \Sigma_c)$ where $\Sigma_c$ is the covariance matrix of $\{\tilde{x}_n^{(c)}\}_{n=1}^{N_c}$. Here the linear version of FDA that assumes $\Sigma_c=I$ for all $c\in [C]$ is utilized for simplicity. By Bayes theorem, given a sample $\tilde{x}_n$, the score function for label $c$ is acquired by 
%
\begin{equation}\label{eq:score-function}
\delta_c(\hat{x}_n) = \hat{x}_n\tran UU\tran\mu_c - \frac{1}{2}\mu_c\tran UU\tran \mu_c +\log q_c
\end{equation}
which is a linear equation in terms of $\hat{x}_n$. In Eqn.(\ref{eq:score-function}), $q_c$ is the prior probability of the $c$-th class, which is estimated by $q_c=N_c/N$. We normalize $\delta_c(\hat{x}_n),\,c=1\cdots C$ with softmax function to obtain the final class prediction probability
\begin{equation}\label{eq:fda-propbability}
p(y_n|\hat{x}_n) = \frac{\exp ^{\delta_{y_n}(\hat{x}_n)}}{\sum_{c=1}^C\exp ^{\delta_{c}(\hat{x}_n)}}
\end{equation}
Substituting Eqn.(\ref{eq:fda-propbability}) into Eqn.(\ref{eq:metric-score}) gives us the transferability metric score of Reg-FDA.


\subsection{Self-Challenging Mechanism by Noise Augmentation}\label{sec:self-challenging}
Though Reg-FDA can project static features for better classes separability,  pre-trained models perform differently on hard examples during fine-tuning (Fig.\ref{fig:fig3}(d)). 
We propose a self-challenging mechanism augmented by the proposed Confidential Mix (ConfMix) noise to encourage pre-trained models to discriminate on hard examples. 
The self-challenging framework consists of a two-stage Reg-FDA. 
In the first stage, the Reg-FDA provides a confidential probability of correctly classifying each sample through Eqn.(\ref{eq:fda-propbability}). 
In the second stage, the Reg-FDA challenges the classification accuracy in the first stage by ConfMix. 

\noindent\textbf{Formulation of ConfMix.} We denote $p(y_n|\hat{x}_n)$ in Eqn.(\ref{eq:fda-propbability}) as $p_n$. Since $p_n$ represents the confidential probability of correctly classifying $n$-th sample in target dataset, a smaller $p_n$ indicates larger difficulty in classifying $(\hat{x}_n,y_n)$. To increase the difficulty of classification, ConfMix builds Reg-FDA in the second stage on the convex combination of the underlying sample and the mean of its outer classes by $p_n$, as written by
\begin{equation}\label{eq:confmix}
\bar{x}_n = p_n \hat{x}_n + (1-p_n) \mu_{c\neq y_n}
\end{equation}
where $\mu_{c\neq y_n} = \frac{1}{N-N_{y_n}} \sum_{n=1, y_n\neq c}^N \hat{x}_n$ denotes the outer classes mean relative to the $c$-th class. 

\begin{figure}[t]
\centering
\includegraphics[scale=0.47]{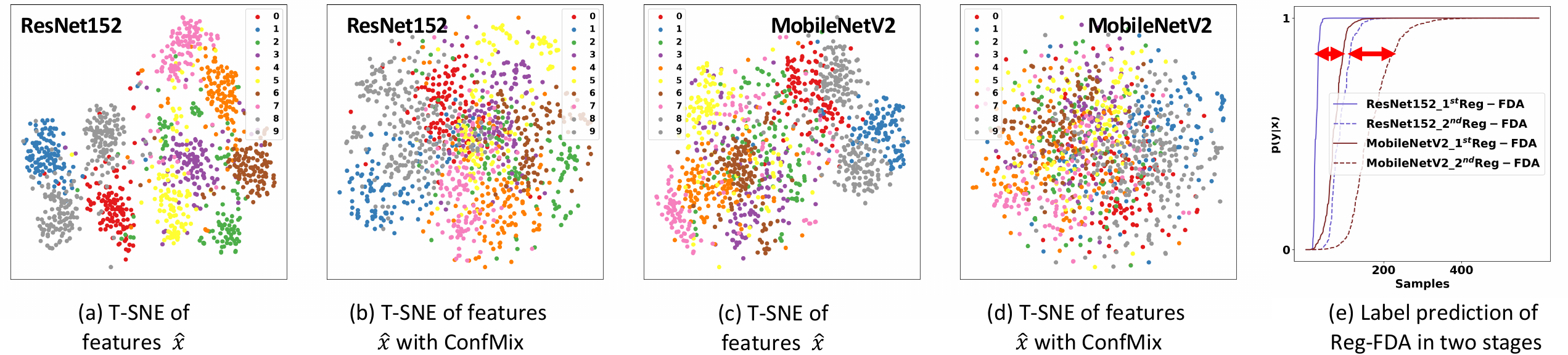}
\caption{(a-d) show that our proposed ConfMix noise reduces separability between classes on both ResNet-152 and MobileNetV2. (e) shows that ConfMix increases the classification difficulty and encourages pre-trained models to differentiate on hard examples. The results are obtained on a fraction of the CIFAR-10 dataset.}
\label{fig:fig4}
\end{figure}

\noindent\textbf{Analysis of ConfMix.} We show that ConfMix increases classification difficulty and improves the discrimination on hard examples for pre-trained models. Through Eqn.(\ref{eq:confmix}),  $\hat{x}_n$ is moved to its outer classes by ConfMix. Hence, different classes would be closer to each other, increasing the difficulty in separating classes by Reg-FDA. For example, Fig.\ref{fig:fig4} show that ConfMix encourages separate feature distribution between classes. Moreover, the final classification probability $p(y_n|\bar{x}_n)$ obtained by Reg-FDA in the second stage is lower than $p(y_n|\hat{x}_n)$ in the first stage, which means the classification by Reg-FDA becomes more difficult. In addition, we also observe from Fig.\ref{fig:fig3}(e \& f) and Fig.\ref{fig:fig4}(e) that the difference between pre-trained models on hard examples are enlarged.

\subsection{Extension to Top-$k$ Model Ensembles Selection}
Top-$k$ model ensembles selection utilizes $k$ pre-trained models to perform ensemble transfer learning. The $k$ models are selected to obtain the best performance on target tasks. The main challenge of top-$k$ model selection is the heterogeneous feature representations extracted from different pre-trained models. For example, ResNet-50 outputs a feature with the dimension of $2048$ while the features extracted from ResNet-18 are $512$-D. Such heterogeneity makes it difficult to compare different pre-trained models.
To simplify the problem, recent works propose to select $k$ models by top-$k$ ranked transferability metrics. 
Despite the simplicity, it fails to consider the relationship between models. In other words, transferability metrics can only measure how well a single model performs on a target task.
However, the complementarity between models should also be considered in ensemble transfer learning \cite{chang2015complementarity,nguyen2021complementary}.

\noindent\textbf{Homogeneous features by SFDA.} Fortunately, our proposed SFDA can deal with heterogeneous feature representations by the projection of $U$. By SFDA, we have $\tilde{x} = U\tran\hat{x} \in \mathbb{R}^{D'}$ where $D'= \min \{D, C-1\}$. Note that $\hat{x}\in\mathbb{R}^D$ where $D$ varies for different models. Fortunately, the projected features of different models by SFDA have homogeneous dimension of $D'=C-1$. Here we assume $D \geq C-1$ which is a common case in practice.

\noindent\textbf{Complementarity score.} By SFDA, we can collect an ensemble of homogeneous features on a model basis given a sample, as denoted by $F^{\ens}=[\tilde{x}^1, \cdots, \tilde{x}^M] \in \mathbb{R}^{M\times D'}$. Hence, each model is now represented by each row of $F^{\ens}$. An ablative approach \cite{zhou2021domain} is employed to evaluate how a model is complementary to other models, as given by
\begin{equation}\label{eq:complementarity-score}
T^{\com}_m = \|F^{\ens}\|_* - \|F^{\ens} \odot 1_m\|_*
\end{equation}
where $T^{\com}_m$ denotes the complementarity score of the $m$-th model,  $1_m\mathbb{R}^M$ is a mask vector with $m$-th entry of $0$ and $1$ elsewhere. In Eqn.(\ref{eq:complementarity-score}), $\|\cdot\|_*$ denotes nuclear norm which is a relaxation of rank of a matrix 
Therefore $T^{\com}_m$ measures how the $m$-th model influences the rank of $F^{\ens}$. A larger $T^{com}_m$ implies that the $m$-th model is more important for the complementarity in $F^{\ens}$. 
Evaluation of the complementarity score is illustrated in Algorithm 2 of Appendix Sec.A.4, where the final score is averaged over a fraction of target samples. 

\noindent\textbf{Ensemble score.} The total ensemble score for selecting top-$k$ models is determined by combining SFDA and complementarity scores. The former evaluates the transferability of a single model, and the latter measures the complementarity between models. Putting them together yields
\begin{equation}\label{eq:ensem-score}
T^{\ens}_m  = r T^{\SFDA}_m + (1-r) T^{\com}_m 
\end{equation}
where $T^{\SFDA}_m$ and $T^{\com}_m$ denotes the transferability score and complementarity score. $r$ is a weight ratio to balance two scores. Empirically, we set $r=0.5$. We select top-$k$ ranked models by $T^{\ens}_m$ to perform ensemble transfer learning.

\section{Experiments}\label{sec:experiment}
This section evaluates our method SFDA on different categories of pre-trained models, including both supervised and self-supervised Convolutional Neural Network (CNN) models and recently-popular vision transformer models \cite{dosovitskiy2020image,shao2021dynamic} (see Appendix B.2). Moreover, we also show the effectiveness of SFDA in the top-$k$ ensembles selection of pre-trained models. The evaluation is performed on $11$ standard classification benchmark in transfer learning. Finally, we conduct an ablation study to analyze our SFDA (also see Appendix B.4).


\subsection{Benchmarks}\label{sec:dataset-gt}

\textbf{Target datasets.} 
We adopt $11$ classification benchmarks widely used in the transfer learning study, including FGVC Aircraft \cite{maji2013fine}, Caltech-101 \cite{fei2004learning}, Stanford Cars \cite{krause2013collecting}, CIFAR-10~\cite{krizhevsky2009learning}, CIFAR-100~\cite{krizhevsky2009learning}, DTD~\cite{cimpoi2014describing}, Oxford 102 Flowers~\cite{nilsback2008automated}, Food-101~\cite{bossard2014food}, Oxford-IIIT Pets~\cite{parkhi2012cats}, SUN397~\cite{xiao2010sun}, and VOC2007~\cite{everingham2010pascal}.
These datasets cover a broad range of classification tasks, which includes scene, texture,
and coarse/fine-grained image classification. 

\noindent\textbf{Ground truth.} 
We can obtain the ground-truth ranking by fine-tuning all pre-trained models with hyper-parameters sweeping on target datasets.
%
Here we consider the normal training setting which is more general in practice than other sparse training methods \cite{zhou2021effective}. Details of target datasets and fine-tuning schemes are described in Appendix Sec.B.1

\subsection{Evaluation on Supervised CNN Models}\label{sec:sup-CNN}

\textbf{Models.} We first evaluate the performance of transferability metrics on ranking pre-trained supervised CNN models. We select $11$ widely-used models including ResNet-34~\cite{he2016deep}, ResNet-50~\cite{he2016deep}, ResNet-101~\cite{he2016deep}, ResNet-152~\cite{he2016deep}, DenseNet-121~\cite{huang2017densely}, DenseNet-169~\cite{huang2017densely}, DenseNet-201~\cite{huang2017densely}, MNet-A1~\cite{tan2019mnasnet}, MobileNetV2~\cite{sandler2018mobilenetv2}, GoogleNet~\cite{szegedy2015going}, and InceptionV3~\cite{szegedy2016rethinking}. 
All these models are trained on ImageNet dataset\cite{krizhevsky2012imagenet}.
We fine-tune these models on the $11$ target datasets to obtain the ground truth (original results are shown in Appendix Sec.B.1).
In addition, the transferability scores of metrics including LEEP, NLEEP, LogME and our SFDA are also calculated 
by following the paradigm in Fig.2.
We re-implement these algorithms in our framework for a fair comparison.

\begingroup
\renewcommand{\arraystretch}{1.1}
\setlength{\tabcolsep}{4pt}
\begin{table}[t]
\begin{center}
\caption{Comparison of different transferability metrics on supervised CNN models in rank correlation $\tau_w$ with the ground truth and the wall-clock time. A larger $\tau_w$ indicates the better transferability assessment of pre-trained models. The best results are denoted in bold. A shorter wall-clock time suggests that the transferability metric is more efficient to obtain. Our proposed SFDA achieves the best transferability prediction over $11$ target tasks while being much more efficient than NLEEP. }
\label{tab:sup-tw}
\scalebox{0.82}{
\begin{tabular}{c c c c c c c c c c c c}
\hline\noalign{\smallskip}
 & Aircraft & Caltech & Cars & CF-10 & CF-100 & DTD & Flowers & Food & Pets & SUN & VOC\\
\noalign{\smallskip}
\hline
\noalign{\smallskip}
\multicolumn{12}{c}{Weighted Kendall's tau $\tau_w$} \\
\noalign{\smallskip}
\hline
LEEP  & -0.234& 0.605& 0.367& 0.824& 0.677& 0.486& -0.243& 0.491& 0.389& 0.701& 0.446\\
LogME  & 0.506 & 0.435 & \textbf{0.576} & 0.852& 0.692& 0.647& 0.111& 0.385& 0.411& 0.511& 0.478\\
NLEEP  & 0.495& 0.661& 0.265 & 0.806 & 0.823& \textbf{0.777}& 0.215& 0.624 & 0.599& \textbf{0.807}&0.654  \\
\underline{SFDA}  & \textbf{0.615} & \textbf{0.737} & 0.487& \textbf{0.949} & \textbf{0.866}& {0.597}& \textbf{0.542}& \textbf{0.815}& \textbf{0.734}& {0.703}& \textbf{0.763} \\
\hline
\noalign{\smallskip}
\multicolumn{12}{c}{Wall-Clock Time (s)} \\
\noalign{\smallskip}
\hline
LEEP  & 5.1& 4.9& 8.3& 22.3& 23.8& 3.5& 3.8 & 37.1 & 3.9& 21.1& 4.8\\
LogME  & 11.5 & 11.0 & 18.3 & 23.5& 36.9& 8.7& 10.9& 54.0& 9.4& 51.8& 9.7\\
NLEEP  & 253.8& 488.7& 973.8 & 1.1e4 & 1.7e4 & 146.0 & 294.0& 2.0e4 & 580.8 & 8.6e3 & 678.8  \\
\underline{SFDA}  & 91.9 & 246.2 & 274.6& 576.6 & 617.9& 171.9& 167.2& 703.7& 181.2& 560.5& 75.0 \\
\hline
\end{tabular}
}
\end{center}
\end{table}
\endgroup

\noindent\textbf{Performance Comparison.}  
We assess transferability metrics by weighted Ken-dall's tau $\tau_w$ that measures the rank correlation between ground truth and metrics scores. 
We compare SFDA with previous LEEP, LogME, and NLEEP. As shown in Table \ref{tab:sup-tw}, SFDA achieves the best rank correlation $\tau_w$ with the ground truth on $8$ target datasets. For example, SFDA outperforms NLEEP by $0.327$, $0.191$, and $0.160$ rank correlation $\tau_w$ on Flowers, Food, and Pets, respectively. On these three datasets, the relative improvements are $152.1$\%, $30.6$\%, and $26.7$\%, respectively, showing the effectiveness of our SFDA in measuring the transferability of pre-trained models. 
On the other hand, for the remaining $3$ target datasets (i.e. Cars, DTD, and SUN397),
our SFDA still has a marginal gap compared to the best-performing transferability metric.

\noindent\textbf{Wall-clock time comparison.} We also provide wall-clock time comparison in Table \ref{tab:sup-tw}. The wall-clock time measures how long it takes for each metric to calculate the transferability scores of all models on a target dataset. 
We can see that both LEEP and LogME are fast to obtain on all target tasks. However, LEEP and LogME are not stable in measuring the transferability score of pre-trained models for all target tasks.
For example, the rank correlation $\tau_w$ of LogME and LEEP on multiple target datasets such as Flowers, Food, and Pets are lower than $0.5$, implying that they are ineffective in measuring transferability on these downstream tasks.
Moreover, NLEEP performs well on a large number of target tasks. But the computation of NLEEP on some target datasets costs even several hours, which is comparable with the fine-tuning procedure. 

As we can see from Table \ref{tab:sup-tw}, our SFDA can measure transferability of models well while only requiring several hundred seconds for computation on all target tasks, achieving a better trade-off between assessment of transferability and computation consumption than other methods.

\begingroup
\renewcommand{\arraystretch}{1.1}
\setlength{\tabcolsep}{4pt}
\begin{table}[t]
\begin{center}
\caption{Comparison of different transferability metrics on self-supervised CNN models regarding $\tau_w$ and the wall-clock time.
Our proposed SFDA achieves the best transferability assessment over $11$ target tasks and exhibits higher efficiency than NLEEP. }
\label{tab:selfsup-tw}
\scalebox{0.82}{
\begin{tabular}{c c c c c c c c c c c c}
\hline\noalign{\smallskip}
 & Aircraft & Caltech & Cars & CF-10 & CF-100 & DTD & Flowers & Food & Pets & SUN & VOC\\
\noalign{\smallskip}
\hline
\noalign{\smallskip}
\multicolumn{12}{c}{Weighted Kendall's tau $\tau_w$} \\
\noalign{\smallskip}
\hline
LogME  & 0.223 & 0.051 & 0.375 & 0.295& -0.008& 0.627& 0.604& 0.570& 0.684& 0.217& 0.158\\
NLEEP  & -0.029& \textbf{0.525} & {0.486} & -0.044 & 0.276& 0.641& 0.534& 0.574 & \textbf{0.792}& 0.719& -0.101  \\
\underline{SFDA}  & \textbf{0.254} & {0.523} & \textbf{0.515}& \textbf{0.619} & \textbf{0.548}& \textbf{0.749}& \textbf{0.773}& \textbf{0.685}& {0.586}& {0.698}& \textbf{0.568} \\
\hline
\noalign{\smallskip}
\multicolumn{12}{c}{Wall-Clock Time (s)} \\
\noalign{\smallskip}
\hline
LogME  & 33.5 & 33.0 & 72.7 & 89.3& 116.0& 15.0& 41.0& 140.1& 40.9& 112.5 &16.7\\
NLEEP  & 581.7& 787.7& 1.6e3 & 8.0e3 & 2.1e4 & 332.7 & 322.3& 1.1e4 & 186.5 & 3.9e4 & 678.8  \\
\underline{SFDA}  & 300.7 & 316.4 & 553.4& 504.1 & 753.1& 170.2& 335.1& 980.7& 157.7& 992.5& 134.7 \\
\hline
\end{tabular}
}
\end{center}
\end{table}
\endgroup

\subsection{Evaluation on Self-Supervised CNN Models}\label{sec:self-sup-CNN}

\textbf{Models.} 
In transfer learning, self-supervised pre-trained models generally have better transferability than their supervised counterpart~\cite{he2020momentum,ericsson2021well}. 
Moreover, the different self-supervised algorithms would provide diverse feature representation for a downstream task~\cite{ericsson2021well}.
Hence, it is essential to investigate selecting the best self-supervised pre-trained model for a target task. 
To this end, we build a pool of self-supervised pre-trained models with ResNet-50 \cite{he2016deep}, including BYOL \cite{grill2020bootstrap}, Deepclusterv2 \cite{caron2018deep}, Infomin \cite{tian2020makes}, MoCov1 \cite{he2020momentum}, MoCov2 \cite{chen2020improved}, Instance Discrimination~\cite{wu2018unsupervised}, PCLv1~\cite{li2020prototypical}, PCLv2~\cite{li2020prototypical}, Selav2~\cite{asano2019self}, SimCLRv1~\cite{chen2020simple}, SimCLRv2~\cite{chen2020big}, and SWAV~\cite{cimpoi2014describing}.
Our goal is to rank these self-supervised models on $11$ downstream tasks in Sec.\ref{sec:dataset-gt}.

\noindent\textbf{Performance Comparison.} We compare our SFDA with LogME and NLEEP on transferability assessment in rank correlation $\tau_w$ with the ground truth. 
Because self-supervised models usually do not have a classifier, LEEP is not included for comparison as it relies on a classifier to calculate transferability score. 
As shown in Table \ref{tab:selfsup-tw}, SFDA still performs consistently well in measuring the transferability of self-supervised models.
On the contrary, LogME and NLEEP fail on some target tasks. For example, LogME and NLEEP have a negative of small positive rank correlation on Aircraft and CIFAR-100. Averaging $\tau_w$ over $11$ target tasks,  the improvement of SFDA ($0.593$) over LogME ($0.345$) and NLEEP ($0.397$) are $71.9$\% and $49.4$\%, respectively. 
The results demonstrate the effectiveness of our SFDA in transferability assessment.

\noindent\textbf{Wall-clock time comparison.}  
As shown in Table \ref{tab:selfsup-tw}, LogME and SFDA usually take tens of seconds to calculate transferability score on all target tasks, while NLEEP may cost several hours on some target datasets.
Though LogME is fast, it is not stable, as mentioned above.
Instead, our SFDA adapts to different pre-trained models and requires several hundred seconds on all target tasks. 
Hence, SFDA is still effective and efficient in evaluating the transferability of self-supervised models.

\subsection{Extension to Top-$k$ Model Ensembles Selection}\label{sec:extend-topk}

\begingroup
\renewcommand{\arraystretch}{1.1}
\setlength{\tabcolsep}{4pt}
\begin{table}[t]
\begin{center}
\caption{Separate effects of components in SFDA. Three variants of SFDA are considered: (1) LogME + ConfMix; (2) SFDA w/o ConfMix; (3) SFDA with $\lambda=0.5$ in Eqn.(\ref{eq:lambda}). We see that both Reg-FDA and the self-challenging mechanism achieved by ConfMix are crucial to SFDA.}
\label{tab:sup-tw-ablation}
\scalebox{0.82}{
\begin{tabular}{c c c c c c c c c c c c}
\hline\noalign{\smallskip}
Variants & Aircraft & Caltech & Cars & CF-10 & CF-100 & DTD & Flowers & Food & Pets & SUN & VOC\\
\noalign{\smallskip}
\hline
(1)  & 0.408 & 0.324 & 0.365 & 0.924& 0.571& 0.328& 0.023& 0.466& 0.390& 0.419& 0.695\\
(2)  & 0.481& 0.676 & 0.403 & 0.887 & 0.773& 0.471& \textbf{0.668}& 0.812 & 0.652& 0.606& 0.721  \\
(3)  &0.424& 0.627 & 0.178 & 0.931 & 0.828& 0.458& 0.228& 0.802 & 0.409& 0.651& 0.650  \\
\underline{SFDA}  & \textbf{0.615} & \textbf{0.737} & \textbf{0.487}& \textbf{0.949} & \textbf{0.866}& \textbf{0.597}& {0.542}& \textbf{0.815}& \textbf{0.734}& \textbf{0.703}& \textbf{0.763}\\
\hline
\end{tabular}
}
\end{center}
\end{table}
\endgroup

\begingroup
\renewcommand{\arraystretch}{1.1}
\setlength{\tabcolsep}{4pt}
\begin{table}[t]
\begin{center}
\caption{Comparison between SFDA and re-training head in terms of $\tau_w$. Our proposed SFDA generally performs better than re-training head over $11$ target tasks. }
\label{tab:retrain-head-tw}
\scalebox{0.82}{
\begin{tabular}{c c c c c c c c c c c c}
\hline\noalign{\smallskip}
 & Aircraft & Caltech & Cars & CF-10 & CF-100 & DTD & Flowers & Food & Pets & SUN & VOC\\
\noalign{\smallskip}
\hline
\noalign{\smallskip}
\multicolumn{12}{c}{Weighted Kendall's tau $\tau_w$} \\
\noalign{\smallskip}
\hline
Re-Head & -0.008& \textbf{0.590} & \textbf{0.666} & 0.583 & 0.501& 0.721& 0.661& \textbf{0.787} & 0.\textbf{703}& 0.637& 0.490  \\
\underline{SFDA}  & \textbf{0.254} & {0.523} & {0.515}& \textbf{0.619} & \textbf{0.548}& \textbf{0.749}& \textbf{0.773}& {0.685}& {0.586}& \textbf{0.698}& \textbf{0.568} \\
\hline
\noalign{\smallskip}
\multicolumn{12}{c}{Wall-Clock Time (s)} \\
\noalign{\smallskip}
\hline
Fine-tune & 3.3e5& 2.8e5 & 2.7e5 & 2.5e5 & 2.5e5& 3.1e5& 3.8e5& 2.6e5 & 2.9e5& 3.9e5& 2.9e5  \\
Re-Head & 2211& 2198 & 2246 & 2219 & 2215& 2210& 2173& 2211 & 2232& 2228 &2375  \\
\underline{SFDA}  & 300.7 & 316.4 & 553.4& 504.1 & 753.1& 170.2& 335.1& 980.7& 157.7& 992.5& 134.7\\
\hline
\end{tabular}
}
\end{center}
\end{table}
\endgroup

\begingroup
\renewcommand{\arraystretch}{1.1}
\setlength{\tabcolsep}{4pt}
\begin{table}[t]
\begin{center}
\caption{Comparison on all CNN models in Sec.\ref{sec:sup-CNN} and Sec.\ref{sec:self-sup-CNN}. The weighted Kendall's tau $\tau_w$ is used to assess transferability metrics. Our SFDA is still better at measuring the transferability of pre-trained models than LogME and NLEEP.}
\label{tab:supall-tw-ablation}
\scalebox{0.82}{
\begin{tabular}{c c c c c c c c c c c c}
\hline\noalign{\smallskip}
Method & Aircraft & Caltech & Cars & CF-10 & CF-100 & DTD & Flowers & Food & Pets & SUN & VOC\\
\noalign{\smallskip}
\hline
LogME  & 0.168 & 0.033 & \textbf{0.506} & 0.687& 0.507& 0.580& 0.301& 0.535& 0.629& 0.284& 0.531\\
NLEEP  & -0.026& 0.714 & 0.099 & 0.491 & 0.653& \textbf{0.766}& 0.373& 0.233 & {0.768}& \textbf{0.716}& 0.637  \\
\underline{SFDA}  & \textbf{0.350}& \textbf{0.791} & 0.450 & \textbf{0.748} & \textbf{0.803}& 0.544& \textbf{0.741}& \textbf{0.743} & \textbf{0.788}& 0.537& \textbf{0.798}  \\
\hline
\end{tabular}
}
\end{center}
\end{table}
\endgroup

Our proposed SFDA is competent for the problem of top-$k$ model ensembles selection as  SFDA makes features extracted from different models homogeneous. Due to the homogeneity, the complementarity between models is considered by SFDA through Eqn.(\ref{eq:complementarity-score}), which we denoted as SFDA$^{\com}$ as shown in Table \ref{tab:sup-ensem-tw}. To verify the effectiveness of SFDA$^{\com}$ in top-$k$ model ensembles selection, we compare it with baselines \cite{agostinelli2021transferability,you2021ranking} that select $k$ models by top-$k$ ranked LogME, NLEEP. We also experimented on selecting $k$ models by top-$k$ ranked SFDA. When $k$ models are selected, we perform ensemble fine-tuning on $11$ target tasks respectively by following the paradigm in \cite{agostinelli2021transferability}. The final ensemble fine-tuning accuracy is used to compare different ensemble transferability metrics. Wee can see in Appendix B.3 that SFDA$^{\com}$ leads to higher fine-tuning accuracy on most target tasks than other metrics.



\subsection{Ablation Analysis}

\textbf{The effect of Reg-FDA and self-challenging mechanism.} SFDA consists of a Reg-FDA module and a self-challenging mechanism achieved by ConfMix. Here we study their separate effects on transferability assessment. To this end, we evaluate the transferability of supervised CNN models in Sec.\ref{sec:sup-CNN}  with the following variants of SFDA, (1) SFDA with Reg-FDA replaced by LogME because LogME can also output a label prediction probability; (2) SFDA w/o ConfMix; and (3) SFDA with a fixed regularization coefficient $\lambda=0.5$ in Eqn.(\ref{eq:Reg-FDA}). 
The results are reported in Table \ref{tab:sup-tw-ablation}. From (1), we see that `LogME + ConfMix' performs worse than SFDA (`Reg-FDA + ConfMix'), indicating that the ConfMix cooperates better with Reg-FDA than LogME. The main reason is that our Reg-FDA projects features for better class separability, and ConfMix increases the difficulty in separating classes features. But LogME has nothing to do with classes separability. Moreover, comparing (2) and SFDA, we conclude that self-mechanism achieved by ConfMix can consistently improve the performance on $11$ downstream tasks. Lastly, an adaptive regularization term in Eqn.(\ref{eq:lambda}) helps SFDA deal with various feature distributions. Fixing the regularization strength leads to worse transferability assessment as shown in Table \ref{tab:sup-tw-ablation} (3).


\noindent\textbf{Comparison with re-training Head.} 
Re-training head is a widely-adopted tool to measure how well the features extracted from the pre-trained model can predict their labels by a classification head. 
It freezes the extracted features and trains the classification head only. After training, the head produces the labels of features. Hence, we can obtain re-training head accuracy. 
Though re-training head simplifies the fine-tuning process, it is still built on static representations and requires a grid search for hyper-parameters such as learning rate and regularization strength.
Hence, re-training head is less efficient and effective than our SFDA. Table \ref{tab:retrain-head-tw} shows that SFDA is more effective in measuring the transferability of pre-trained models than the re-training head. In particular, our SFDA is
more than 600x faster than brute-force fine-tuning in running time.

\noindent\textbf{Performance on total pre-trained model hubs.} In practice, we may have a large-scale pre-trained model hubs rather than categorized ones in Sec.\ref{sec:sup-CNN} - \ref{sec:self-sup-CNN}. 
In this case, we need to rank a variety of pre-trained models. To this end, we consolidate
the pre-trained models from supervised CNN models in Sec.\ref{sec:sup-CNN} and self-supervised CNN models in Sec.\ref{sec:self-sup-CNN} into one group (including $23$ pre-trained models in total), then applying the ranking methods on it. The results are shown in Table \ref{tab:supall-tw-ablation}. The
results further reveal that our SFDA performs consistently well on a larger group of pre-trained models compared to LogME and NLEEP.

\section{Conclusions and Discussions}
The rapid development of deep learning produces many deep models pre-trained on different source datasets with various learning strategies. Given numerous pre-trained models, it is practical to consider how to rank them and screen the best ones for target tasks. In this work, we answer this question by proposing a new transferability metric named Self-challenging Fisher Discriminant Analysis (SFDA). SFDA build upon a regularized Fisher Discriminate Analysis and a self-challenging mechanism.
Compared with prior arts, SFDA behaves more like fine-tuning in terms of classes separability and discrimination on hard examples, characterizing itself as an efficient, effective and robust transferability assessment method.
Moreover, SFDA can be naturally extended to multiple pre-trained model ensemble selection where complementarity between models is essential for better ensemble performance on downstream tasks. 
SFDA measures transferability by mimicking the fine-tuning procedure for downstream classification tasks. 

\noindent\textbf{Discussions.} Although SFDA is fast and effective in measuring transferability of pre-trained models, it can only used in downstream classification tasks. Several future works are worth investigating in pursuit of a more universal metric. Firstly, an interesting future work would be how to extend SFDA in other types of target tasks such as regression tasks by characterizing the features of fine-tuning on these tasks. Secondly, investigating SFDA in out-of-distribution setting \cite{yang2021generalized,zhou2022model} could also be a fruitful future direction.

\noindent\textbf{Acknowledgement.} We thank anonymous reviewers from the venue ECCV 2022 for their valuable comments. We also thank Xiuzhe Wu and Ruichen Luo for their helpful discussions. Ping Luo is supported by the General Research Fund of HK No.27208720, No.17212120, and No.17200622.

\clearpage
%
%
\bibliographystyle{splncs04}
\bibliography{egbib}

\clearpage

\begin{center}
\large{\textbf{Appendix}}
\end{center}
The appendix provides more details about the main paper in both methods in Sec.\ref{sec:app-method} and experiments in Sec.\ref{sec:app-exp}.
\renewcommand\thesection{\Alph{section}}
\setcounter{section}{0}

\section{More Details about SFDA}\label{sec:app-method}

\subsection{Interpretation of weighted Kendall's tau}
The Kendall's $\tau$ represents the ratio of concordant pairs minus discordant pairs when enumerating all 
 \scalebox{0.7}{$\begin{pmatrix} M \\ 2 \end{pmatrix}$} pairs of $\{T_m\}_{m=1}^M$ and $\{G_m\}_{m=1}^M$ as given by
\begin{equation}
\tau = \frac{2}{M(M-1)}\sum_{1\leq i < j \leq M}\mathrm{sgn}(G_i-G_j)\mathrm{sgn}(T_i-T_j)
\end{equation}
where $\mathrm{sgn}(x)$ is a sign function returning $1$ if $x>0$ and $-1$ otherwise.
Moreover, we use a weighted version of Kendall's $\tau$, denoted as $\tau_w$, to evaluate transferability metrics considering that a best performing pre-trained model is always preferred for target task in transfer learning. $\tau_w$ can measure the ranking performance of top performing models. In principle, a larger $\tau_w$ indicates the transferability metric can produce a better ranking for pre-trained models.

\subsection{Algorithm of SFDA}
\begin{algorithm}
	\caption{Pipeline of SFDA.}
	\label{alg:SFDA}
	{\fontsize{9}{9} \selectfont
		\begin{algorithmic}[1]
			\State {\bfseries Input:} target dataset \scalebox{0.9}{
			$\mathcal{T}=\{(x_n, y_n)\}_{n=1}^N$}; $M$ pre-trained models \scalebox{0.9}{
			$\{\phi_m\}_{m=1}^M$};\\
			{\bfseries Hyper-parameters:} $a=4$ in Eqn.(\ref{eq:lambda}) to generate the regularization coefficient. \\
			{\bfseries Output:} the transferability scores of SFDA, \scalebox{0.9}{
			$\{T_m\}_{m=1}^M$};.
			%
			%
			%
			%
			%
			\State {\bfseries Set:} self-challenge  = True 
			\For{$m=1$ {\bfseries to} $M$}
			\State calculate: static representations \scalebox{0.8}[0.9]{$\hat{x} = \theta_m(x)$}.
			\State split: \scalebox{0.9}{\(\mathcal{T}=\{\hat{x}_n^{(1)}\}_{n=1}^{N_1} \cup \cdots \cup \{\hat{x}_n^{(C)}\}_{n=1}^{N_C}\)}.
		    \State calculate: \scalebox{0.9}{\(\mu=\sum_{n=1}^N\hat{x}_n ,
\mu_c=\sum_{n=1}^{N_c}\hat{x}_n^{(c)}\)}. \Comment{total mean and class mean.}
			\State calculate: \scalebox{0.8}{\(S_W=\sum_{c=1}^{C}\sum_{n=1}^{N_c}(\hat{x}_n^{(c)}-\mu_c)(\hat{x}_n^{(c)}-\mu_c)\tran\)}. \Comment{within scatter.}
			\State calculate: \scalebox{0.8}{\( S_B= \sum_{c=1}^CN_c(\mu_c - \mu)(\mu_c - \mu)\tran\)}. \Comment{between scatter.}
			\State calculate: \scalebox{0.8}{\(\lambda =1 / (1+ \exp ^{-a\sigma(S_B)})\)}. \Comment{regularization strength.}
			\State solve $U$: \scalebox{0.8}{\(S_B U =[\mathrm{tr}(U\tran S_B U)/\mathrm{tr}(U\tran \tilde{S}_W U)] \tilde{S}_W U\)} in Eqn.(\ref{eq:solution-FDA}). 
			\State calculate: updated representations \scalebox{0.8}[0.9]{$\tilde{x} = U\tran \hat{x}$} in Fisher Space.
			\State calculate: \scalebox{0.8}[0.9]{$\delta_c(\hat{x}_n) = \hat{x}_n\tran UU\tran\mu_c - \frac{1}{2}\mu_c\tran UU\tran \mu_c +\log q_c$}.
			\State calculate: \scalebox{0.8}[0.9]{\(p(y_n|x_n) =  \exp ^{\delta_{y_n}(\hat{x}_n)}/{\sum_{c=1}^C\exp ^{\delta_{c}(\hat{x}_n)}}\)}
			\If {self-challenge}
				\State calculate: \scalebox{0.8}[0.9]{\(\hat{x}_n = p_n \hat{x}_n + (1-p_n) \mu_{c\neq y_n}\)}\Comment{ConfMix Noise.}
				\State {\bfseries Set:} self-challenge  = False
				\State turn to line 7. 
			
			\EndIf
			
			\State Calculate: \scalebox{0.9}{\(T_m = \frac{1}{N}\sum_{n=1}^N\log p(y_n|x_n)\)}. \Comment{Transferability score of SFDA.}
			\EndFor
		\end{algorithmic}
	}
\end{algorithm}

Algorithm \ref{alg:SFDA} illustrates the whole pipeline of our proposed SFDA. As we can see, SFDA consists of a two-stage Reg-FDA. The Reg-FDA in the second stage challenges the Reg-FDA in the first stage by the proposed ConfMix noise. SFDA has an obvious advantage compared with previous transferability metrics such as LogME, because it explicitly projects static features $\{\hat{x}_n\}_{n=1}^N$ into a Fisher space where features exhibits better linear separability as shown in line 13 in Algorithm 1. Besides, the updated features $\{\hat{x}_n\}_{n=1}^N$ becomes more discriminative in classification difficulty, behaving more like finetuning than other metrics. Moreover, SFDA is fast to obtain, as it requires no gradient optimization and only involves generalized eigenvalue problem. 

Note that the largest eigenvalue of $\sigma(S_B)$ in line $11$ of Algorithm \ref{alg:SFDA} can be obtained by iteration method \cite{miyato2018spectral}. Specifically, we calculate  $\sigma(S_B)$ by the following iteration. For $s=1,2\cdots S$,
\begin{equation}\label{eq:sigma_SB}
v_s = S_B\tran u_{s-1} /\|S_B\tran u_{s-1}\|_2\,\,\mathrm{and}\,\, u_s = S_B\tran v_{s} /\|S_B\tran v_{s}\|_2
\end{equation}
where we initialize $u_0$ as a vector of all ones. After $S$ iterations, we have $\sigma(S_B)=u_S\tran S_B v_S$. In practice, we find that $S=3$ is enough for obtaining a precise $\sigma(S_B)$. Note that Eqn.(\ref{eq:sigma_SB}) only involves matrix-vector product. $\sigma(S_B)$ can be efficiently acquired.

\subsection{Feature visualization by t-SNE}
\begin{figure}[t]
\centering
\includegraphics[scale=0.46]{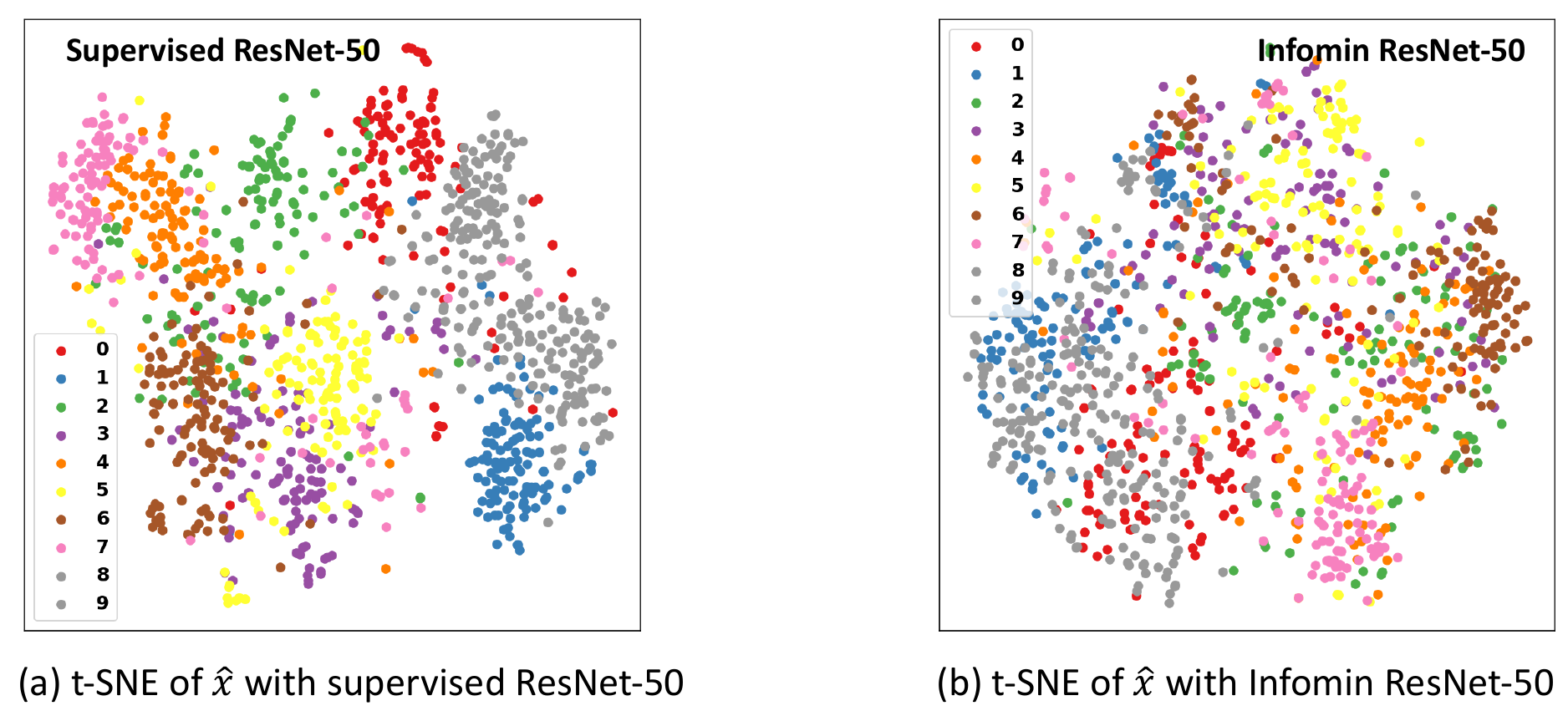}
\caption{(a \& b) show that Infomin ResNet-50 has a larger within scatter of classes than its supervised counterpart on CIFAR-10 dataset.}
\label{fig:fig5}
\end{figure}
In Sec.\ref{sec:Reg-FDA}, we treat $\lambda\in [0,1]$ as an adaptive regularization strength,  considering the diverse distribution of features $\{\hat{x}_i\}_{i=1}^N$ extracted from different pre-trained models. For example, as shown in Fig.\ref{fig:fig5}, supervised ResNet-50 has a larger between scatter of classes than its self-supervised counterpart with Infomin on CIFAR-10 dataset, implying that ResNet-50 with Infomin needs stronger supervision on minimizing within scatter of every class for better classes separation.

\subsection{Algorithm of SFDA for top-$k$ model ensembles selection}

\begin{algorithm}
	\caption{Top-$k$ model ensembles of SFDA.}
	\label{alg:topk-SFDA}
	{\fontsize{9}{9} \selectfont
		\begin{algorithmic}[1]
			\State {\bfseries Input:} projected features in Fisher space \scalebox{0.9}{
			$\{(\tilde{x}_n^m\}_{n=1}^{N_{\ens}}, m=1,2,\cdots M$} where the superscript $m$ denotes the $m$-th model and $N_{\ens}$ is the number of samples used to select top-$k$ models;\\
			{\bfseries Hyper-parameters:} $k$ to the number of selected models, $r=0.5$ in Eqn.(\ref{eq:ensem-score}). \\
			{\bfseries Output:} top-$k$ models ensemble;.
			%
			%
			%
			%
			%
			\For{$n=1$ {\bfseries to} $N_{\ens}$}
			\State calculate: models feature ensemble $F^{\ens}_n=[\tilde{x}^1_n, \cdots, \tilde{x}^M_n]$.
			\State calculate: complementarity score $(T^{\com}_m)_n = \|F^{\ens}_n\|_* - \|F^{\ens}_n \odot 1_m\|_*$
		    \State calculate: model ensemble score $(T^{\ens}_m)_n   = r (T^{\SFDA}_m)_n + (1-r) (T^{\com}_m)_n $.
			\EndFor
		\State calculate: $T^{\ens}_m$ by averaging $(T^{\ens}_m)_n$ under all $N_{ens}$ input samples.
		\State ranking: select $k$ models by top-$k$ ranked ensemble $T^{\ens}_m, n=1,2,\cdots M$		
		\end{algorithmic}
	}
\end{algorithm}

Here we provide the framework of SFDA for top-$k$ model ensembles selection in Algorithm \ref{alg:topk-SFDA}. As we can see, the total ensemble score for selecting top-$k$ models is determined by combining SFDA and complementarity scores. The former evaluates the transferability of a single model, and the latter measures the complementarity between models. In experiment, we set $N_{\ens}=3000$ which is large enough to measure the complementarity score precisely and efficiently (tens second to run Algorithm \ref{alg:topk-SFDA} for each target task).

\section{More Experimental Results}\label{sec:app-exp}

\subsection{Results of ground truth of fine-tuning}
\textbf{Fine-tuning details.} The ground-truth of the problem of pre-trained models ranking is to fine-tune all pre-trained models with a hyper-parameters sweep on target datasets.
Given the model and the target dataset, two of the most important parameters would be learning rate and weight decay in optimizing the model \cite{li2020rethinking}. Therefore, we careful fine-tune pre-trained models with a grid search of learning rate in $\{1e-1, 1e-2, 1e-3, 1e-4\}$ and weight decay in $\{1e-3,1e-4, 1e-5, 1e-6, 0\}$. After determining the best hyper-parameters candidate, we fine-tune the pre-trained model on the target dataset with the candidate and then obtain the test accuracy as the ground truth. We use a Tesla V100 with a batch size of $128$ to perform finetuning. All input images are resized to $224\times 224$. To avoid random error, we repeat the above fine-tuning procedure three times and take an average to obtain the final fine-tuning accuracy. For reference, we list the fine-tuning accuracy of supervised CNN models in Sec.\ref{sec:sup-CNN}, self-supervised CNN models in Sec.\ref{sec:self-sup-CNN}, and vision transformer models in Sec.\ref{sec:vit} in Table \ref{tab:fine-tuning-supcnn}, Table \ref{tab:fine-tuning-selfsupcnn}, and Table \ref{tab:fine-tuning-vit}, respectively. To obtain ensemble finetuning accuracy, we also use the above hyper-parameters sweep. To avoid huge memory consumption, we firstly finetune each pre-trained model on target dataset and then fix the static representation extracted by the fine-tuned pretrained model. After that, we train $k$ classification head for each model, and average the resulting class logits to make the final label prediction.

\noindent\textbf{Hardware for counting wall-clock time.} For all wall-clock time counting, we use Intel(R) Xeon(R) Platinum CPU.

\begingroup
\renewcommand{\arraystretch}{1.1}
\setlength{\tabcolsep}{4pt}
\begin{table}[t]
    \centering
    \caption{The fine-tuning accuracy of supervised CNN models on $11$ target tasks.}
    \label{tab:fine-tuning-supcnn}
    \scalebox{0.8}{%
    \begin{tabular}{l c c c c c c c c c c c}
	\hline\noalign{\smallskip}
 		& Aircraft & Caltech & Cars & CF-10 & CF-100 & DTD & Flowers & Food & Pets & SUN & VOC\\
	\noalign{\smallskip}
	\hline
ResNet-34& 84.06& 91.15& 88.63& 96.12& 81.94& 72.96& 95.2& 81.99& 93.5& 61.02& 84.6\\
ResNet-50& 84.64& 91.98& 89.09& 96.28& 82.8& 74.72& 96.26& 84.45& 93.88& 63.54& 85.8\\
ResNet-101& 85.53& 92.38& 89.47& 97.39& 84.88& 74.8& 96.53& 85.58& 93.92& 63.76& 85.68\\
ResNet-152& 86.29& 93.1& 89.88& 97.53& 85.66& 76.44& 96.86& 86.28& 94.42& 64.82& 86.32\\
DenseNet-121& 84.66& 91.5& 89.34& 96.45& 82.75& 74.18& 97.02& 84.99& 93.07& 63.26& 85.28\\
DenseNet-169& 84.19& 92.51& 89.02& 96.77& 84.26& 74.72& 97.32& 85.84& 93.62& 64.1& 85.77\\
DenseNet-201& 85.38& 93.14& 89.44& 97.02& 84.88& 76.04& 97.1& 86.71& 94.03& 64.57& 85.67\\
MNet-A1& 66.48& 89.34& 72.58& 92.59& 72.04& 70.12& 95.39& 71.35& 91.08& 56.56& 81.06\\
MobileNetV2& 79.68& 88.64& 86.44& 94.74& 78.11& 71.72& 96.2& 81.12& 91.28& 60.29& 82.8\\
Googlenet& 80.32& 90.85& 87.76& 95.54& 79.84& 72.53& 95.76& 79.3& 91.38& 59.89& 82.58\\
InceptionV3& 80.15& 92.75& 87.74& 96.18& 81.49& 72.85& 95.73& 81.76& 92.14& 59.98& 83.84\\
 \hline      
    \end{tabular}

    }

\end{table}
\endgroup

\begingroup
\renewcommand{\arraystretch}{1.1}
\setlength{\tabcolsep}{4pt}
\begin{table}[t]
    \centering
    \caption{The fine-tuning accuracy of self-supervised CNN models on $11$ target tasks.}
    \label{tab:fine-tuning-selfsupcnn}
    \scalebox{0.8}{%
    \begin{tabular}{l c c c c c c c c c c c}
	\hline\noalign{\smallskip}
 		& Aircraft & Caltech & Cars & CF-10 & CF-100 & DTD & Flowers & Food & Pets & SUN & VOC\\
	\noalign{\smallskip}
	\hline
BYOL& 82.1& 91.9& 89.83& 96.98& 83.86& 76.37& 96.8& 85.44& 91.48& 63.69& 85.13\\
Deepclusterv2& 82.43& 91.16& 90.16& 97.17& 84.84& 77.31& 97.05& 87.24& 90.89& 66.54& 85.38\\
Infomin& 83.78& 80.86& 86.9& 96.72& 70.89& 73.47& 95.81& 78.82& 90.92& 57.67& 81.41\\
InsDis& 79.7& 77.21& 80.21& 93.08& 69.08& 66.4& 93.63& 76.47& 84.58& 51.62& 76.33\\
MoCov1& 81.85& 79.68& 82.19& 94.15& 71.23& 67.36& 94.32& 77.21& 85.26& 53.83& 77.94\\
MoCov2& 83.7& 82.76& 85.55& 96.48& 71.27& 72.56& 95.12& 77.15& 89.06& 56.28& 78.32\\
PCLv1& 82.16& 88.6& 87.15& 96.42& 79.44& 73.28& 95.62& 77.7& 88.93& 58.36& 81.91\\
PCLv2& 83.0& 87.52& 85.56& 96.55& 79.84& 69.3& 95.87& 80.29& 88.72& 58.82& 81.85\\
Sela-v2& 85.42& 90.53& 89.85& 96.85& 84.36& 76.03& 96.22& 86.37& 89.61& 65.74& 85.52\\
SimCLRv1& 80.54& 90.94& 89.98& 97.09& 84.49& 73.97& 95.33& 82.2& 88.53& 63.46& 83.29\\
SimCLRv2& 81.5& 88.58& 88.82& 96.22& 78.91& 74.71& 95.39& 82.23& 89.18& 60.93& 83.08\\
SWAV& 83.04& 89.49& 89.81& 96.81& 83.78& 76.68& 97.11& 87.22& 90.59& 66.1& 85.06\\
 \hline      
    \end{tabular}

    }

\end{table}
\endgroup

\begingroup
\renewcommand{\arraystretch}{1.1}
\setlength{\tabcolsep}{4pt}
\begin{table}[t]
    \centering
    \caption{The fine-tuning accuracy of vision transformer models on $11$ target tasks.}
    \label{tab:fine-tuning-vit}
    \scalebox{0.8}{%
    \begin{tabular}{l c c c c c c c c c c c}
	\hline\noalign{\smallskip}
 		& Aircraft & Caltech & Cars & CF-10 & CF-100 & DTD & Flowers & Food & Pets & SUN & VOC\\
	\noalign{\smallskip}
	\hline
ViT-T& 71.26& 89.39& 82.09& 96.52& 81.58& 71.86& 95.5& 81.96& 91.44& 58.4& 83.1\\
ViT-S& 73.12& 92.7& 86.72& 97.69& 86.62& 75.08& 96.79& 86.26& 94.02& 64.76& 86.62\\
ViT-B& 78.39& 93.47& 89.26& 98.56& 89.96& 77.66& 97.98& 88.96& 94.61& 68.62& 87.88\\
PVTv2-B2& 84.14& 93.13& 90.6& 97.96& 88.24& 77.16& 97.89& 88.67& 93.86& 66.44& 86.44\\
PVT-T& 69.76& 90.04& 84.1& 94.87& 75.26& 72.92& 95.8& 83.78& 91.48& 61.86& 84.6\\
PVT-S& 75.2& 93.02& 87.61& 97.34& 86.2& 75.77& 97.32& 86.98& 94.13& 65.78& 86.62\\
PVT-M& 76.7& 93.75& 87.66& 97.93& 87.36& 77.1& 97.36& 85.56& 94.48& 67.22& 87.36\\
Swin-T& 81.9& 91.9& 88.93& 97.34& 85.97& 77.04& 97.4& 86.67& 94.5& 65.51& 87.54\\
MoCov3-S& 76.04& 89.84& 82.18& 97.92& 85.84& 71.88& 93.89& 82.84& 90.44& 60.6& 81.84\\
DINO-S& 72.18& 86.76& 79.81& 97.96& 85.66& 75.96& 95.96& 85.69& 92.59& 64.14& 84.8\\
 \hline      
    \end{tabular}

    }

\end{table}
\endgroup

\subsection{Evaluation on Vision Transformer Models}\label{sec:vit}

\textbf{Models.} Vision transformer (ViT) models have attracted much attention recently due to its power in processing multi-modal data. When pre-training ViT models, various model architectures and data augmentation settings result in models with drastically different performance. Hence, how to select a model for further adaptation for an end application is significant in practice. We compare SFDA with other metrics in ranking pre-trained ViT models in terms of transferability. To this end, we collect $10$ ViT models including ViT-T \cite{dosovitskiy2020image}, ViT-S \cite{dosovitskiy2020image}, ViT-B \cite{dosovitskiy2020image}, DINO-S \cite{caron2021emerging}, MoCov3-S \cite{chen2021empirical} , PVTv2-B2 \cite{wang2021pyramid}, PVT-T  \cite{wang2021pyramid}, PVT-S  \cite{wang2021pyramid}, PVT-M  \cite{wang2021pyramid}, and Swin-T \cite{liu2021swin}. The ground truth of models' transferability are obtained by fine-tuning these models on $11$ downstream tasks as shown in Table \ref{tab:fine-tuning-vit}.
\begingroup
\renewcommand{\arraystretch}{1.1}
\setlength{\tabcolsep}{4pt}
\begin{table}[t]
\begin{center}
\caption{Comparison of different transferability metrics on ViT models in terms of $\tau_w$ and the wall-clock time. We see that our proposed SFDA achieves better trade-off between transferability assessment and computation consumption over $11$ target tasks. }
\label{tab:vit-tw}
\vspace{-0.2in}
\scalebox{0.82}{
\begin{tabular}{c c c c c c c c c c c c}
\hline\noalign{\smallskip}
 & Aircraft & Caltech & Cars & CF-10 & CF-100 & DTD & Flowers & Food & Pets & SUN & VOC\\
\noalign{\smallskip}
\hline
\noalign{\smallskip}
\multicolumn{12}{c}{Weighted Kendall's tau $\tau_w$} \\
\noalign{\smallskip}
\hline
LogME  & 0.299 & 0.382 & 0.642 & 0.741& \textbf{0.723}& 0.569& {0.512}& 0.580& 0.528& 0.619& 0.519\\
NLEEP  & -0.282& 0.027 & \textbf{0.693} & 0.674 & 0.538& 0.123& -0.262& 0.105 & 0.409& 0.268& 0.109  \\
\underline{SFDA}  & \textbf{0.533} & \textbf{0.533} & 0.632& \textbf{0.743} & {0.692}& \textbf{0.570}& \textbf{0.515}& \textbf{0.592}& \textbf{0.787}& \textbf{0.707}& \textbf{0.809} \\
\hline
\noalign{\smallskip}
\multicolumn{12}{c}{Wall-Clock Time (s)} \\
\noalign{\smallskip}
\hline
LogME  & 5.7 & 3.6 & 11.2 & 13.1& 21.9& 14.2& 3.6& 33.1& 4.9& 186.1 &3.6\\
NLEEP  & 553.7& 716.8& 1.1e3 & 8.0e3 & 1.2e4 & 183.7 & 819.2& 3.4e4 & 256.4 & 2.7e4 & 288.3  \\
\underline{SFDA}  & 31.3 & 34.0 & 59.1& 121.7 & 140.3& 20.7& 28.7& 218.8& 45.6& 129.8& 26.5 \\
\hline
\end{tabular}
}
\end{center}
\end{table}
\endgroup

\noindent\textbf{Performance Comparison.} We compare our SFDA with LogME and NLEEP on transferability assessment in terms of rank correlation $\tau_w$. The results are reported in Table \ref{tab:vit-tw}, SFDA still performs consistently well in measuring transferability of self-supervised models. An interesting observation is that LogME outperform NLEEP on evaluation of ViT models, which is not the case on evaluation supervised and self-supervised CNN models. We guess that LogME is adept in deal with sequential feature representation extracted from ViT models as LogME is also designed for regression downstream tasks.
Averaging $\tau_w$ over $11$ target tasks,   SFDA ($0.647$) improves rank correlation $\tau_w$ by $196.8$\% and $16.3$\% relative to NLEEP ($0.218$) and LogME ($0.556$), respectively. Hence, our SFDA can measure the transferability of pre-trained ViT models better.

\noindent\textbf{Wall-clock time comparison.}  We provide wall-clock time comparison in Table \ref{tab:vit-tw}. We can see that LogME is efficient enough to calculate transferability score on all target tasks.  Moreover, NLEEP performs worse than LogME and our SFDA in terms of both rank correlation $\tau_w$ and computation efficiency. In addition, our SFDA is much more efficient in computing transferability score while achieving the best transferability assessment. 

\subsection{Results on Top-$k$ Model Ensembles Selection}
\textbf{Performance Comparison.} Table \ref{tab:sup-ensem-tw} shows that SFDA$^{\com}$ leads to higher fine-tuning accuracy on most target tasks. It demonstrates SFDA's superiority to those metrics which do not consider the complementarity between models. For example, when performing top-$2$ and top-$3$ ensembles selection, SFDA$^{\com}$ outperforms LogME, NLEEP, and SFDA on $9$ and $8$ downstream tasks. Therefore, SFDA$^{\com}$ is effective in multiple pre-trained model ensembles selection.

\begingroup
\renewcommand{\arraystretch}{1.1}
\setlength{\tabcolsep}{4pt}
\begin{table}[t]
\begin{center}
\caption{Comparison of different transferability metrics on top-$k$ model ensembles selection. Results are obtained by selecting top-$2$ and top-$3$ models among supervised CNN models and them perform ensemble fine-tuning following \cite{agostinelli2021transferability}. {SFDA} and {SFDA$^{\com}$} select top-$k$ models by top-$k$ ranked SFDA scores and top-$k$ ranked ensemble score through Eqn.(\ref{eq:ensem-score}), respectively.
SFDA$^{\com}$ generally performs well over $11$ target tasks as it considers complementarity between models. }
\label{tab:sup-ensem-tw}
\vspace{-0.1in}
\scalebox{0.75}{
\begin{tabular}{c c c c c c c c c c c c c}
\hline\noalign{\smallskip}
Top-$k$ & Method & Aircraft & Caltech & Cars & CF-10 & CF-100 & DTD & Flowers & Food & Pets & SUN & VOC\\
\noalign{\smallskip}
\hline
\multirow{4}{*}{$k=2$} 
&LogME  & 87.23 & \textbf{93.87} & 91.38 & 97.86& \textbf{86.98}& 77.14& 96.87& 87.43& \textbf{94.62}& 65.47& 86.46\\
&NLEEP  & 85.17 & 93.45 & 91.03 & 97.86 & \textbf{86.98}& \textbf{77.25}& 97.04& 86.62 & 94.59& \textbf{65.81} & \textbf{86.64}  \\
&{SFDA}  &\textbf{87.71} &93.40 &91.03 &97.86 &\textbf{86.98} & 76.95 & \textbf{97.59} & \textbf{87.95} & 94.58 & \textbf{65.81} & 86.46\\
&\underline{SFDA$^{\com}$}  &\textbf{87.71} &\textbf{93.87}	 &\textbf{91.69} &\textbf{97.91} &\textbf{86.98} & 76.95 & {97.24} & \textbf{87.95} & \textbf{94.62} & \textbf{65.81} & 86.46\\
\hline
\multirow{4}{*}{$k=3$} 
&LogME  & 87.23 & 93.87 & 91.80 & 97.88& 86.96& 77.68& 97.56& 87.83& 94.70& 66.20& \textbf{86.89}\\
&NLEEP  & 86.98 & \textbf{94.12} & 91.76 & \textbf{98.02} & \textbf{87.48}& \textbf{78.14}& 97.65& 87.37 & 94.71& \textbf{66.95}& \textbf{86.89}  \\
&{SFDA}  &\textbf{88.01} &93.95 &91.76 &\textbf{98.02} &\textbf{87.48} & 77.68 & \textbf{97.99} & \textbf{88.49} & 94.82 & 66.53 & \textbf{86.89}\\
&\underline{SFDA$^{\com}$}  &\textbf{88.01} &93.87 &\textbf{91.95} &\textbf{98.02} &\textbf{87.48} & \textbf{78.14} & 97.35 & \textbf{88.49} & \textbf{94.92} & 66.53 & \textbf{86.89}\\
\hline
\end{tabular}
}

\end{center}
\end{table}
\endgroup

\subsection{More ablation study}
\noindent\textbf{SFDA under other transferability assessment measures.} Other than weighted Kendall's tau, here we also adopt different types of measurement to evaluate our SFDA. The measures include Kendall's tau ($\tau$), Pearson's correlation ($r$), weighted Pearson's correlation ($r_w$), top-$k$ relative accuracy denoted as Rel@$k$ that is the ratio between the best fine-tuning accuracy on the downstream task with the top-$k$ ranked models and the best fine-tuning precision with all the models. We test the robustness of transferability metrics to different measurements using supervised CNN models on the CIFAR-10  and CIFAR-100 datasets in Table \ref{tab:sup-measurements}. Our SFDA consistently outperforms previous transferability metrics such as LEEP, LogME and NLEEP under the above measurements, showing the superiority of SFDA.

\begingroup
\renewcommand{\arraystretch}{1.1}
\setlength{\tabcolsep}{4pt}
\begin{table}[t]
\begin{center}
\caption{SFDA under different measurements of transferability assessment on CIFAR-10 and CIFAR-100 datasets using supervised CNN models.}
\label{tab:sup-measurements}
\vspace{-0.2in}
\scalebox{0.65}{
\begin{tabular}{|c| c c c c c c c| c| c c c c c c c|}
\hline\noalign{\smallskip}
Data &Method & Rel@$1$ & Rel@$3$ & $r$ &  $r_w$ & $\tau$ &  $\tau_w$ &Data &Method & Rel@$1$ & Rel@$3$ & $r$ &  $r_w$ & $\tau$ &  $\tau_w$ \\
\noalign{\smallskip}
\cline{1-16}
\multirow{4}{*}{CF10}&LEEP  & \textbf{1.0} & \textbf{1.0} & 0.623 & 0.753& 0.673& 0.824&\multirow{4}{*}{CF100} &LEEP  & 0.991 & \textbf{1.0} & 0.653 & 0.692& 0.624& 0.677\\
&LogME  & \textbf{1.0}& \textbf{1.0} & 0.718 & 0.756 & 0.782& 0.852& &LogME  & \textbf{1.0} & \textbf{1.0} & 0.508 & 0.586& 0.477& 0.692\\
&NLEEP  & \textbf{1.0}& \textbf{1.0} & 0.635 & 0.774 & 0.636& 0.806& &NLEEP  & \textbf{1.0} & \textbf{1.0} & \textbf{0.694} & 0.762& 0.734& 0.823\\
&\underline{SFDA}  & \textbf{1.0} & \textbf{1.0} & \textbf{0.768}& \textbf{0.801} & \textbf{0.891}& \textbf{0.949} & &\underline{SFDA}  & \textbf{1.0} & \textbf{1.0} & {0.691} & \textbf{0.764}& \textbf{0.771}& \textbf{0.896}\\
\hline
\end{tabular}
}
\vspace{-0.1in}
\end{center}
\end{table}
\endgroup


\end{document}